\documentclass[10pt]{article} 
\usepackage[accepted]{tmlr}
\newcommand{\openreview}{\url{https://openreview.net/forum?id=bYItLUU135}}
\usepackage{titlesec}
\titlespacing*{\subsection}{0pt}{0.5em}{0.5em}

\usepackage{amsmath,amsfonts,bm}

\def\eqref#1{equation~\ref{#1}}

\def\1{\bm{1}}

\DeclareMathAlphabet{\mathsfit}{\encodingdefault}{\sfdefault}{m}{sl}
\SetMathAlphabet{\mathsfit}{bold}{\encodingdefault}{\sfdefault}{bx}{n}

\usepackage[utf8]{inputenc} 
\usepackage[T1]{fontenc}    
\usepackage{hyperref}       
\usepackage{url}            
\usepackage{booktabs}       
\usepackage{amsfonts}       
\usepackage{nicefrac}       
\usepackage{microtype}      
\usepackage{xcolor}  
\usepackage{microtype}
\usepackage{graphicx}
\usepackage{subfigure}
\usepackage{booktabs} 
\usepackage{wrapfig}
\usepackage{hyperref}
\usepackage{booktabs}

\usepackage{siunitx}
\usepackage{soul}

\usepackage{amsmath}
\usepackage{amssymb}
\usepackage{mathtools}
\usepackage{amsthm}
\usepackage{hyperref}
\usepackage{url}
\usepackage[utf8]{inputenc} 
\usepackage[T1]{fontenc}    
\usepackage{hyperref}       
\usepackage{url}            
\usepackage{booktabs}       
\usepackage{amsfonts}       
\usepackage{hyperref}
\usepackage{url}
\usepackage[utf8]{inputenc}
\usepackage{adjustbox}
\usepackage{amsmath, bm} 
\usepackage{amssymb}  
\usepackage{upgreek}
\usepackage{booktabs}
\usepackage{subcaption}
\usepackage{lipsum}
\usepackage[utf8]{inputenc}
\newtheorem{theorem}{Theorem}[section]
\usepackage{svg}
\usepackage{multicol}

\usepackage[acronyms,nonumberlist,nopostdot,nomain,nogroupskip,acronymlists={hidden}]{glossaries} 
\newglossary[algh]{hidden}{acrh}{acnh}{Hidden Acronyms}

\usepackage[capitalize,noabbrev]{cleveref}

\theoremstyle{plain}

\newtheorem{lemma}[theorem]{Lemma}

\theoremstyle{definition}

\theoremstyle{remark}
\newtheorem{remark}[theorem]{Remark}
\newcommand{\dataset}{{\cal D}}

\hypersetup{hidelinks}
\usepackage[normalem]{ulem}
\usepackage{xcolor}

\usepackage{mdframed}
\newmdenv[
  linecolor=red,
  linewidth=1pt,
  backgroundcolor=red!5,
  frametitle={\color{red}\sout{Deleted}},
]{deletedblock}

\newtheorem{condition}{Condition}

\usepackage[textsize=tiny]{todonotes}
\usepackage[ruled,lined,noend]{algorithm2e}
\usepackage{float}

\title{Recursive Deep  Inverse Reinforcement Learning}
\newacronym{dl}{DL}{Deep Learning}
\newacronym{irl}{IRL}{Inverse Reinforcement Learning}
\newacronym{gcl}{GCL}{Guided Cost Learning}
\newacronym{airl}{AIRL}{Adversarial Inverse Reinforcement Learning}
\newacronym{gail}{GAIL}{Generative Adversarial Imitation Learning}
\newacronym{mlirl}{ML-IRL}{Maximum Likelihood Inverse Reinforcement Learning}
\newacronym{ub}{UB}{Upper Bound}
\newacronym{ekf}{EKF}{Extended Kalman Filter}
\newacronym{ioc}{IOC}{Inverse Optimal Control}
\newacronym{map}{MAP}{Maximum A-Posteriori}
\newacronym{dnn}{DNN}{Deep Neural Network}
\newacronym{sgd}{SGD}{Stochastic Gradient Descent}
\newacronym{rdirl}{RDIRL}{Recursive Deep Inverse Reinforcement Learning}
\newacronym{kf}{KF}{Kalman Filter}
\newacronym{mppi}{MPPI}{Model Predictive Path Integral Control}
\newacronym{fim}{FIM}{Fisher Information Matrix}
\newacronym{mlp}{MLP}{Multi-Layer Perceptron}

 \author{%
   Paul Ghanem \\
   Department of Computer Engineering\\
   Northeastern University\\
   Boston, MA 02115 \\
   \texttt{ghanem.p@northeastern.edu} \\
   \AND
    Owen Howell \\
   Boston Dynamics AI Institute\\
   Boston, MA 02142 \\
   \texttt{howell.o@northeastern.edu} \\
    \AND
    Michael Potter \\
   Department of Computer Engineering\\
   Northeastern University\\
   Boston, MA 02115 \\
   \texttt{potter.m@northeastern.edu} \\
    \AND
    Pau Closas \\
   Department of Computer Engineering\\
   Northeastern University\\
   Boston, MA 02115 \\
   \texttt{pau.closas@northeastern.edu} \\
   \AND
    Alireza Ramezani \\
   Department of Computer Engineering\\
   Northeastern University\\
   Boston, MA 02115 \\
   \texttt{a.ramezani@northeastern.edu} \\
   \AND
    Deniz Erdogmus \\
   Department of Computer Engineering\\
   Northeastern University\\
   Boston, MA 02115 \\
   \texttt{D.erdogmus@northeastern.edu} \\
   \AND
    Tales Imbiriba \\
   Department of Computer Engineering\\
   University of Massachusetts\\
   Boston, MA 02125 \\
   \texttt{tales.imbiriba@umb.edu} \\   
 }

\begin{document}

\maketitle

\begin{abstract}
Inferring an adversary's goals from exhibited behavior is crucial for counterplanning and non-cooperative multi-agent systems in domains like cybersecurity, military, and strategy games. Deep \gls{irl} methods based on maximum entropy principles show promise in recovering adversaries' goals but are typically offline, require large batch sizes with gradient descent, and rely on first-order updates, limiting their applicability in real-time scenarios. We propose an online \gls{rdirl} approach to recover the cost function governing the adversary actions and goals. 
Specifically, we minimize an upper bound on the standard \gls{gcl} objective using sequential second-order Newton updates, akin to the \gls{ekf}, leading to a fast (in terms of convergence) learning algorithm. We demonstrate that \gls{rdirl} is able to recover cost and reward functions of expert agents in standard and adversarial benchmark tasks. Experiments on benchmark tasks show that our proposed approach outperforms several leading IRL algorithms.
\end{abstract}
\glsresetall

\section{Introduction}

\gls{ioc} and \gls{irl} aim to infer parameterized cost and reward functions in optimal control and reinforcement learning problems, respectively, from observed state-control data. This data is assumed to be generated by an expert following an optimal policy that either minimizes a cost function or maximizes a reward function.

Previous \gls{irl} approaches have included maximum-margin approaches \citep{abbeel2004apprenticeship}, and probabilistic approaches such as \citet{ziebart2008maximum}. In this work, we build on the maximum entropy \gls{irl} framework presented previously \citep{ziebart2008maximum}. In this framework, training consists of two nested loops. The inner loop approximates the optimal control policy for a hypothesized cost function, while the outer loop minimizes a negative log-likelihood cost function ~\citep{ziebart2008maximum}, constructed by sampling a full trajectory from the inner loop's optimal control policy and by using the expert trajectory that is observed from the expert.

Due to this nested structure, training under the maximum entropy deep \gls{irl} in an online fashion becomes very challenging since inner and outer loops need long trajectories and large batch sizes to converge. Available \gls{irl} approaches exploit the fact that it is often feasible to store and process entire state and control sequences in batches \citep{molloy2018online}. In real-time settings with memory, latency and compute constraints, this is generally not feasible.

Recursive optimization strategies such as \gls{ekf} sequentially minimize a loss function that is a summation of mean square error of observed and estimated states, and mean squared error of the estimated states and their predicted values produced by assumed model dynamics ~\citep{humpherys2012fresh,ghanem2023fast}. Hence, \gls{ekf} cannot be naively leveraged to optimize the negative log-likelihood function ~\citep{ziebart2008maximum} since the log of summation term could not be optimized sequentially. Recent works have proposed moment-matching approaches ~\citep{swamy2021moments,zeng2022maximum,zeng2025structural}, 
leading to objective functions that have a simple summation form, making them more suitable for online adaptive learning. However, they are not explicitly derived from maximum entropy \gls{irl}, and prior formulations have not been optimized in an online setting.

To overcome this limitation, we require a reformulation of the maximum entropy objective into a structure amenable to recursive optimization. To address this gap, we show that the moment matching loss function introduced in ~\citet{swamy2021moments} provides an upper bound for the negative log-likelihood objective of maximum entropy \gls{irl} ~\citep{finn2016guided,ziebart2008maximum}. We then propose a recursive optimization algorithm that minimizes the moment matching loss using expert demonstrations and sampled trajectories from the inner optimal control policy. This approach alleviates the need to optimize the negative log-likelihood cost function only after collecting all trajectories from the inner-loop policy and the expert. Instead, it enables incremental optimization, processing each expert observation as it arrives.

The main contribution of this work is a deep maximum entropy online \gls{irl} algorithm, \gls{rdirl}, that learns nonlinear cost and reward functions parameterized by neural networks directly from expert demonstrations as they arrive. Unlike previous deep learning approaches, our method updates the inner control policy after each new expert sample, enabling online adaptation of policies. By processing state–action pairs sequentially, without storing or batching entire trajectories, \gls{rdirl} is well-suited for real-time applications with memory and latency constraints. Moreover, because the policy and cost updates occur incrementally, our approach converges significantly faster than competing \gls{irl} methods. We validate our approach in simulated benchmark tasks, demonstrating that it outperforms leading \gls{irl} methods.

\section {Related Work}

\gls{irl}, also known as \gls{ioc} \citep{finn2016guided}, aims to learn reward or cost functions from expert agents operating under optimal control or reinforcement learning policies. Several \gls{ioc} methods have been developed to recover finite-horizon optimal control cost functions, including approaches based on Karush-Kuhn-Tucker (KKT) conditions \citep{zhang2019inverse,zhang2019inverse1,puydupin2012convex}, Pontryagin’s minimum principle ~\citep{molloy2022inverse,molloy2020online,jin2020pontryagin}, and the Hamilton-Jacobi-Bellman equation ~\citep{pauwels2014inverse,hatz2012estimating}.

These methods typically follow a two-stage process: first, a feedback gain matrix is computed from state and control sequences using system identification techniques, and second, linear matrix inequalities are solved to recover the objective-function parameters from the feedback gain matrix. Online variations of \gls{ioc} methods based on the Hamilton-Jacobi-Bellman equation ~\citep{zhao2024extended,molloy2018online,molloy2020online,self2020online,self2020online1,self2020online2} have also been developed. However, both offline and online versions of these methods are generally limited to simple parameter estimation, assume partial knowledge of the expert’s cost function, and do not incorporate deep neural network (\gls{dnn}) representations of cost functions.

\gls{irl} approaches have also been proposed based on maximum margin ~\citep{abbeel2004apprenticeship,ratliff2006maximum} and maximum entropy ~\citep{ziebart2008maximum,boularias2011relative}. Among these, maximum entropy \gls{irl}, as introduced by \citet{ziebart2008maximum}, has become one of the leading approaches. In this framework, optimization seeks to find reward or cost function parameters that maximize the likelihood of the observed expert trajectory under a maximum entropy distribution. This involves estimating a partition function from samples drawn from a background distribution that represents a control policy ~\citep{finn2016connection,fu2017learning}, which is dependent on a parameterized cost function. The control policy may range from reinforcement learning ~\citep{ho2016generative,fu2017learning} to receding horizon optimal control ~\citep{xu2022receding}.

Building on maximum entropy \gls{irl}, feature-based methods ~\citep{hadfield2016cooperative,wu2020efficient} model the reward function as an inner product between a feature vector $f$ and a parameter vector $\theta$. These methods have been successfully implemented, with the feature characteristics and parameter vector size typically chosen to match the true cost function structure. However, they assume some structural knowledge of the expert’s cost function or domain knowledge ~\citep{finn2016guided}. Online versions of feature-based maximum entropy \gls{irl} have also been developed ~\citep{rhinehart2018first,arora2021i2rl}, but they have not yet been extended to include a \gls{dnn} parameterization of the reward and cost functions.

Similarly, maximum entropy \gls{irl} with deep learning representations of the reward function has been successfully implemented ~\citep{wulfmeier2015maximum}. These methods, which leverage \glspl{dnn} for complex reward functions, have gained popularity and become widely used ~\citep{finn2016guided,wulfmeier2015maximum,ho2016generative,xu2019learning,xu2022receding,fu2017learning,fu2019language,yu2019meta}. As a result, they have emerged as leading \gls{irl} approaches, outperforming feature-based methods ~\citep{finn2016guided,xu2022receding,ho2016generative}.

In this work, we propose a new online \gls{irl} method based on the maximum entropy framework ~\citep{ziebart2008maximum,ziebart2010modeling}. Unlike other online approaches ~\citep{molloy2018online,self2020online,self2020online2,molloy2020online,rhinehart2018first,arora2021i2rl}, the proposed methodology allows the cost and reward functions to be parameterized using deep neural networks. Our approach is mostly related to the algorithm introduced by \citet{finn2016connection}, which minimizes a negative log-likelihood function and uses \gls{mppi} ~\citep{xu2022receding} as the inner control policy. However, unlike prior work, we recursively adapt the sampling distribution representing the inner control policy each time an expert demonstration is observed.

To summarize, our proposed method is the first to combine several key features into a single effective algorithm. It can learn adversarial cost functions online, which is critical for applications such as evasion and pursuit. Additionally, it can learn complex, expressive cost functions, parameterized by deep neural networks, eliminating the need for manual design of cost functions typically required in recursive methods ~\citep{molloy2018online,zhao2024extended,self2020online}. While some prior methods have demonstrated good performance with online \gls{ioc} \citep{zhao2024extended,molloy2020online,self2020online} and deep neural network-based cost functions \citep{finn2016connection,fu2017learning,ho2016generative,zeng2022maximum,swamy2021moments}, to the best of our knowledge, no previous approach has successfully combined these two properties.

\section{Background}
\subsection{Maximum Entropy Inverse Reinforcement Learning}
Our Inverse reinforcement learning method builds on Guided Cost learning framework \cite{finn2016guided} which is derived from maximum entropy Inverse Reinforcement Learning (IRL) \citep{ziebart2008maximum}. Our method seeks to learn an expert cost function or rewards function by observing the expert's behavior. The framework assumes the demonstrated expert behavior to be the result of the expert acting stochastically and near-optimally with respect to an unknown cost function. Specifically, the model assumes that the expert samples the demonstrated trajectories $\tau_{i}$ from the distribution ~\citep{finn2016guided}: 

\begin{align}
    p_{\theta}(\tau) = \frac{1}{\mathcal{Z}_\theta} \exp(-c_\theta(\tau)) \label{eqn:gcl}
\end{align}

where $\tau = \{x_1, u_1, \ldots, x_N, u_N\}$ is a trajectory sample, $x_N$ and $u_N$ are the agent's observed state and control input at time $N$. $c_\theta(\tau) = \sum_{k=1}^{N} c_\theta(x_k, u_k)$ is the cost function 
parameterized by $\theta \in \mathcal{W} \subseteq \mathbb{R}^{d_\theta}$, 
and $\mathcal{Z}_{\theta} = \int \exp(-c_\theta(\tau))\,d\tau$ is the 
partition function. Here $d_\theta$ denotes the number of parameters and 
$\mathcal{W}$ is the space of admissible parameters determined by the 
chosen network architecture (see Section~B.2). The parametric family 
$\{p_\theta\}_{\theta \in \mathcal{W}}$, equivalently 
$\{c_\theta\}_{\theta \in \mathcal{W}}$, is assumed known and 
twice-differentiable with respect to $\theta$. That is, the functional 
form of $c_\theta$ is fixed a priori and only the parameters $\theta$ 
are unknown and must be inferred from observed expert behavior.

The partition function $\mathcal{Z}$ is difficult to compute for large or continuous domains, and presents the main computational challenge in maximum entropy IRL. In the sample-based approach to maximum entropy IRL  \citep{finn2016guided,fu2017learning,ho2016generative,finn2016connection} the partition function $\mathcal{Z}$ is estimated from a background distribution $q(\tau)$ representing the inner control policy, where $\tau$ are sampled from the policy $q(\tau)$. 
The central idea behind the maximum entropy approach is to estimate $\theta$ that maximizes the likelihood of the maximum entropy distribution $p_{\theta}(\tau)$:
$$
\hat{\theta}= \underset{\theta}{\arg\max} \quad p_{\theta}(\tau).
$$

This approach is equivalent to minimizing the negative log-likelihood 
of Equation \cref{eqn:gcl} given by \citet{finn2016guided}:
\begin{equation}
    \mathcal{L}_{\text{IRL}}(\theta) = \frac{1}{N} \sum_{\tau_i \in 
    \mathcal{D}_{\text{demo}}} c_\theta(\tau_i) + \log \mathcal{Z}_\theta
    \label{eq:nll_exact}
\end{equation}
Since $\mathcal{Z}_\theta$ is intractable to compute 
exactly in large or continuous domains, we resort to importance sampling. 
We introduce a background distribution $q(\tau)$ representing the inner 
control policy, and rewrite the partition function by multiplying and 
dividing the integrand by $q(\tau)$:
\begin{equation}
    \mathcal{Z}_\theta = \int \exp(-c_\theta(\tau))\,d\tau 
    = \int \frac{\exp(-c_\theta(\tau))}{q(\tau)}\,q(\tau)\,d\tau
    = \mathbb{E}_{q(\tau)}\!\left[\frac{\exp(-c_\theta(\tau))}{q(\tau)}\right].
    \label{eq:partition_is}
\end{equation}
This expectation is then approximated by drawing $M$ sample trajectories 
$\{\tau_j\}_{j=1}^{M}$ from $q(\tau)$:
\begin{equation}
    \mathcal{Z}_\theta \approx \frac{1}{M} \sum_{j=1}^{M} 
    \frac{\exp(-c_\theta(\tau_j))}{q(\tau_j)}, 
    \quad \tau_j \sim q(\tau).
    \label{eq:partition_mc}
\end{equation}
Substituting~\eqref{eq:partition_mc} into~\eqref{eq:nll_exact} yields 
the sample-based negative log-likelihood given by \citet{finn2016guided}:
\begin{equation}
    \mathcal{L}_{\text{IRL}}(\theta) \approx \frac{1}{N} 
    \sum_{\tau_i \in \mathcal{D}_{\text{demo}}} c_\theta(\tau_i) + 
    \log \frac{1}{M} \sum_{\tau_j \in \mathcal{D}_{\text{samp}}} 
    \frac{\exp(-c_\theta(\tau_j))}{q(\tau_j)}
    \label{eq:nll_mc}
\end{equation}
where $\mathcal{D}_{\text{samp}}$ is the set of $M$ background samples 
drawn from the inner control policy $q(\tau)$, and 
$\mathcal{D}_{\text{demo}}$ is the set of $N$ expert demonstrations.

To represent the cost function $c_\theta (\tau)$, IOC or IRL feature-based methods typically use a linear combination of hand-crafted features $f:(u, x)\mapsto f(u,x)$, leading to $c_\theta (\tau) = \theta^T f(u_t, x_t)$ \citep{abbeel2004apprenticeship}. This representation is difficult to apply to more
complex domains~\citep{finn2016guided}. 
Recent works have focused on the use of high-dimensional expressive function approximators,  representing $c_\theta (\tau)$ using neural networks, and outperforming feature-based methods ~\citep{finn2016guided,fu2017learning,ho2016generative}. 
In this work, we only leverage neural networks to represent the cost function although, other parameterizations could also be used with our method.
In practice, the negative log-likelihood in \eqref{eq:nll_mc} is minimized using gradient descent and batch training. Previous algorithms using deep networks as the cost function parameterization required long and multiple expert demonstrations and sampled trajectories from background policies in order to converge through multiple training iterations. Moreover, training could not proceed before generating all expert and sampled trajectories which restricted it to offline training paradigms. In this work, we introduce a recursive optimization algorithm that adapts network parameters $\theta$ on the fly whenever an expert demonstration is observed.

\subsection{Kalman Filter as Recursive Second-Order Optimizer}

The \gls{kf} is among the most widely used state estimators in engineering applications.  This algorithm recursively estimates the state variables, for example, the position and velocity of a projectile in a noisy linear dynamical system ~\citep{lipton1998moving}, by minimizing the mean-squared estimation error of the current state, as noisy measurements are received and as the system evolves in time ~\citep{humpherys2012fresh}. Each update provides the latest unbiased estimate of the system variables. Since the updating process is fairly general and relatively easy to compute, the \gls{kf} can often be implemented in real-time. When dealing with nonlinear systems extensions of the \gls{kf} exist such as the \gls{ekf} which resorts to linearizations using first-order Taylor's expansions~\cite{sarkka2023bayesian}.

\citet{humpherys2012fresh} showed that the Kalman filter and the EKF, which are recursive algorithms for estimating states in noisy dynamical systems \citep{lipton1998moving,sarkka2023bayesian}, can be derived as special cases of recursive second-order Newton optimization. While this overarching framework can be applied to loss functions beyond MSE, in the classical filtering context it is applied to cumulative MSE loss functions of the form \citep{humpherys2012fresh}:
\begin{equation}
    J_n(X_n,Y_n) = \sum_{k=1}^{n}j_k(x_k,y_k) 
    \label{eq:sum_cost_KF}
\end{equation}

where $ X_n=\{x_1,\dots,x_n\}$ and $x_n$ represents the state of interest at step $n$. Moreover, $Y_n=\{y_1,\dots,y_n\}$ where $y_n$ represents the measurement data at step $n$. $j_k$ represents the cost at step $k$ associated with $x_k$ and $y_k$, while $J_n$ is the cumulative value of $j_k$ and represents the cumulative cost associated with sequences $X_n$ and $Y_n$.

The \gls{ekf} estimates the state $x_n$ that minimizes \eqref{eq:sum_cost_KF}, where $j_k$ is specifically an MSE cost function, at step $n$ using second-order Newton method as new measurement $y_n$ arrives. Thus, \eqref{eq:sum_cost_KF} can be re-written as: 
\begin{equation}
    J_n(X_n,Y_n) = J_{n-1}(X_{n-1},Y_{n-1}) +  j_n(x_n,y_n) 
    \label{eq:sum_cost_KF_divided}
\end{equation}
The \gls{ekf} finds $x_n$ that minimizes \eqref{eq:sum_cost_KF_divided} given previous loss function $J_{n-1}$, previous state estimates of $X_{n-1}$, previous measurements $Y_{n-1}$ and current measurement $y_n$.

To illustrate, consider a nonlinear system $x_k = f_k(x_{k-1}, u_k) + w_k$ with observations $y_k = h_k(x_k) + v_k$, where $w_k$ and $v_k$ are zero-mean noise with covariances $Q_k$ and $R_k$. The cumulative MSE-EKF loss over the state sequence $X_k = \{x_0, \ldots, x_k\}$ is:
\begin{equation}
    J_k(X_k) = \frac{1}{2}\sum_{i=1}^{k} \lVert y_i - h_i(x_i) 
    \rVert^2_{R_i^{-1}} + \frac{1}{2}\sum_{i=1}^{k} \lVert x_i - 
    f_i(x_{i-1}, u_i) \rVert^2_{Q_i^{-1}}
\end{equation}
Applying a single Newton step with the judicious initial guess described above yields the EKF updates:
\begin{align}
    P_{k|k-1} &= F_k P_{k-1} F_k^\top + Q_k, \quad 
    \hat{x}_{k|k-1} = f_k(\hat{x}_{k-1}, u_k), \\
    P_k &= \left(P_{k|k-1}^{-1} + H_k^\top R_k^{-1} H_k\right)^{-1}, 
    \quad \hat{x}_k = \hat{x}_{k|k-1} - P_k H_k^\top R_k^{-1}
    (h_k(\hat{x}_{k|k-1}) - y_k)
\end{align}
where $F_k = Df_k(\hat{x}_{k-1}, u_k)$ and $H_k = Dh_k(\hat{x}_{k|k-1})$ are the Jacobians, and $P_k$ is the lower-right block of the inverse Hessian (which equals the estimation covariance in the EKF setting).

As shown in \citet{humpherys2012fresh}, when the system dynamics are linear, $J_n$ is a positive-definite quadratic form and a single Newton step yields the exact minimizer regardless of the initial guess. By choosing a judicious initial guess, specifically the minimizer of $J_{n-1}$ extended by one block, the Newton update simplifies to a recursive formula that only requires the previous estimate, the new data, and a matrix $P_n$ that recursively accumulates second-order curvature information. When the dynamics are nonlinear, as in the EKF~\citep{sarkka2023bayesian}, the loss is no longer quadratic and a single Newton step yields only an approximation. 

In classical Kalman filtering applications such as navigation and target tracking ~\citep{ward2006satellite,roumeliotis2000bayesian}, the goal is to estimate states $x_n$ given sequences of noisy (often Gaussian) data $y_n$.
In our IRL setting, however, we aim at estimating the parameters $\theta$ of the cost function $c_{\theta}(\tau)$ from expert demonstration $\tau \in D_\mathrm{demo}$ recursively.
To do this, we apply the recursive second order optimization framework to the moment-matching loss derived in Section~4. Unlike the EKF where each $x_k$ is a distinct state that evolves according to system dynamics, $\theta_k$ is a single fixed unknown and the sequence $\hat{\theta}_1, \hat{\theta}_2, \ldots$ represents successive estimates refined as more expert demonstrations are observed. The per-step cost $j_k$ becomes the moment-matching term plus a regularization penalty. Since the loss is not MSE, the resulting algorithm is not a Kalman filter but is \emph{akin to} the EKF in that each Newton step yields an approximation of parameters $\theta$.The full derivation is presented in Section~5.

\section{Moment Matching as Upper Bound of the Negative Log-likelihood}

In this section, we derive an upper-bound of the negative log-likelihood, leading to an optimization problem that is suitable for \gls{kf}-like online estimation of the parameter vector $\theta$. That is, the resulting upper bound can be written following the same summation structure of equations~\ref{eq:sum_cost_KF} and~\ref{eq:sum_cost_KF_divided}. The log-sum term in \eqref{eq:nll_mc} prevents direct recursive minimization, but the derived upper bound resolves this issue and enables sequential optimization.

We begin by stating two mild, practically motivated conditions that will be needed to establish the bound.

\begin{condition}[Bounded Non-negative Cost Function]\label{cond:cost}
The cost function $c_\theta(\tau)$ is constrained to take bounded values, i.e., there exists a constant $-\infty < c_{\min} < c_{\max} < \infty$ such that:
\begin{equation}
    c_{\min} \leq c_\theta(\tau) \leq c_{\max}, \quad \forall\, \theta \in \mathcal{W}, \; \forall\, \tau.
\end{equation}
This is achievable through standard architectural choices, such as a ReLU activation with output clipping, or a sigmoid output layer.
\end{condition}

\begin{condition}[Bounded Sampling Density]\label{cond:density}
The sampling density $q(\tau)$ is constrained to take non-zero bounded values. 

This is achievable through control policy choice such as MPPI. The MPPI sampling distribution $q(\tau)$ operates over a compact control space $\mathcal{U} = [u_{\min}, u_{\max}]$ (reflecting physical actuator limits) with its mean $\mu$ constrained to $\mathcal{U}$. Since MPPI samples controls from a Gaussian $\mathcal{N}(\mu, \Sigma)$ with $\mu \in \mathcal{U}$, the density evaluated at any $u \in \mathcal{U}$ is bounded:
\begin{equation}
    0 < q_{\min} \leq q(\tau) \leq q_{\max} < \infty, \quad \forall\, \tau \in \mathcal{U},
\end{equation}
where
\begin{align}
    q_{\max} &= (2\pi)^{-d/2} |\Sigma|^{-1/2}, \\
    q_{\min} &= (2\pi)^{-d/2} |\Sigma|^{-1/2} \exp\!\left(-\tfrac{1}{2} D_{\max}\right),
\end{align}
with $D_{\max} = \max_{u \in \mathcal{U},\, \mu \in \mathcal{U}} (u - \mu)^\top \Sigma^{-1} (u - \mu)$ being a finite constant determined by $\mathcal{U}$ and $\Sigma$. Since both $\mu$ and $u$ are constrained to the compact set $\mathcal{U}$ and $\Sigma$ is fixed, both $q_{\min}$ and $q_{\max}$ are finite positive constants determined entirely by algorithmic design parameters ($\mathcal{U}$, $\Sigma$), independent of $\theta$.
\end{condition}

In \citet{matkovic2007variant}, the authors present a Jensen's reversed inequality for convex functions as follows. Let $[a,b]$ be an interval in $\mathbb{R}$, $y_1, \ldots, y_N \in [a,b]$, and $p_1, \ldots , p_N$ be positive real numbers such that $\sum_{n=1}^N p_n = 1$. If $f : [a,b] \rightarrow \mathbb{R}$ is convex on $[a,b]$, then (Corollaries~3 and~4) of \citealp{matkovic2007variant}) :
\begin{equation}
\begin{aligned}
    \sum_{n=1}^{N} p_n f(y_n) - & f \left( \sum_{n=1}^{N} p_n y_n \right) \leq &  f(a) + f(b) - 2f \left(\frac{a+b}{2} \right)
    \label{eqn:reverse_jensen_general}
\end{aligned}
\end{equation}
Replacing the function $f$ by the negative $\log$ function, $f = -\log$ which is a convex function on $]0, \infty [$,  \eqref{eqn:reverse_jensen_general} can be re-written as ~\citet{matkovic2007variant} :

\begin{equation}
\begin{aligned}
\log \left( \sum_{n=1}^{N} p_n y_n \right) & \leq \sum_{n=1}^{N} p_n \log (y_n) -\log (a) - \log(b) + 2 \log \left(\frac{a+b}{2} \right) \label{eqn:reverse_jensen_log}
\end{aligned}
\end{equation}


We apply~\eqref{eqn:reverse_jensen_log} to the log-sum term in the sample-based negative log-likelihood~\eqref{eq:nll_mc}. We define:
\begin{equation}
    p_j = \frac{1}{M} \quad \text{and} \quad y_j = \frac{\exp(-c_\theta(\tau_j^{\mathrm{samp}}))}{q(\tau_j^{\mathrm{samp}})}, \quad j = 1, \ldots, M,
    \label{eqn:pn_yn_def}
\end{equation}
where $\tau_j^{\mathrm{samp}}$ is a trajectory sampled from $\dataset_{\mathrm{samp}}$.

\begin{lemma}[Bounds on $y_j$]\label{lem:bounds}
Under Conditions~\ref{cond:cost} and~\ref{cond:density}, the quantities $y_j$ defined in~\eqref{eqn:pn_yn_def} satisfy $y_j \in [a, b]$ where:
\begin{equation}
    a = \frac{\exp(-c_{\max})}{q_{\max}} > 0, \qquad b = \frac{\exp(-c_{\min})}{q_{\min}} < \infty.
    \label{eqn:ab_bounds}
\end{equation}
Both $a$ and $b$ are positive, finite constants that depend only on architectural and algorithmic design parameters, and are independent of $\theta$.
\end{lemma}

\begin{proof}
Since $c_{\min} \leq c_\theta(\tau) \leq c_{\max}$, we have $\exp(-c_{\max}) \leq \exp(-c_\theta(\tau)) \leq \exp(-c_{\min})$. Combined with $0 < q_{\min} \leq q(\tau) \leq q_{\max}$, we obtain:
\begin{equation}
    \frac{\exp(-c_{\max})}{q_{\max}} \leq \frac{\exp(-c_\theta(\tau))}{q(\tau)} \leq  \frac{\exp(-c_{\min})}{q_{\min}}.
\end{equation}
Note that $a > 0$ follows from the exponential structure of the maximum entropy distribution: since $\exp(-x) > 0$ for all finite $x \in \mathbb{R}$, finiteness of $c_{\max}$ guarantees $\exp(-c_{\max}) > 0$, and thus $a > 0$.
\end{proof}

By Lemma~\ref{lem:bounds}, the requirements of the reverse Jensen's inequality~\eqref{eqn:reverse_jensen_log} are satisfied. Substituting~\eqref{eqn:pn_yn_def} into~\eqref{eqn:reverse_jensen_log}, we obtain an upper bound on the log-sum term:
\begin{equation}
    \log \frac{1}{M} \sum_{j=1}^{M} \frac{\exp(-c_\theta(\tau_j^{\mathrm{samp}}))}{q(\tau_j^{\mathrm{samp}})} 
    \leq \frac{1}{M} \sum_{j=1}^{M} \left(-c_\theta(\tau_j^{\mathrm{samp}}) - \log q(\tau_j^{\mathrm{samp}})\right) - K,
    \label{eqn:log_upper_bound}
\end{equation}
where
\begin{equation}
    K = \log(a) + \log(b) - 2\log\!\left(\frac{a+b}{2}\right),
    \label{eqn:K_def}
\end{equation}
which depends only on $c_{\max}$, $q_{\min}$, and $q_{\max}$, all fixed by the network architecture and MPPI configuration. Therefore, $K$ is independent of $\theta$.

Replacing~\eqref{eqn:log_upper_bound} in the sample-based negative log-likelihood~\eqref{eq:nll_mc}, we derive the following upper bound:
\begin{equation}
\begin{aligned}
    \mathcal{L}_{\text{IRL}}(\theta) 
    &= \frac{1}{N} \sum_{i=1}^{N} c_\theta(\tau_i^{\mathrm{demo}}) + \log \frac{1}{M} \sum_{j=1}^{M} \frac{\exp(-c_\theta(\tau_j^{\mathrm{samp}}))}{q(\tau_j^{\mathrm{samp}})} \\
    &\leq \frac{1}{N} \sum_{i=1}^{N} c_\theta(\tau_i^{\mathrm{demo}}) + \frac{1}{M} \sum_{j=1}^{M} \left(-c_\theta(\tau_j^{\mathrm{samp}}) - \log q(\tau_j^{\mathrm{samp}})\right) - K.
\end{aligned}
\label{eqn:upper_bound_general}
\end{equation}
Since $\log q(\tau_j^{\mathrm{samp}})$ and $K$ are independent of $\theta$---noting that in practice $q(\tau)$ is determined by the cost function parameters from the previous optimization iteration, which are held fixed during the current update---minimizing this upper bound over $\theta$ is equivalent to minimizing:
\begin{equation}
    \mathcal{L}_{\mathrm{UB\text{-}MM}} = \frac{1}{N} \sum_{i=1}^{N} c_\theta(\tau_i^{\mathrm{demo}}) - \frac{1}{M} \sum_{j=1}^{M} c_\theta(\tau_j^{\mathrm{samp}}).
    \label{eqn:moment_matching_loss_general}
\end{equation}
By multiplying and dividing the second term by $N$ and partitioning the $M$ background samples into $N$ groups of $M/N$ samples each (one group per expert demonstration), we can factor out $\frac{1}{N}$ and rewrite~\eqref{eqn:moment_matching_loss_general} as:
\begin{equation}
    \mathcal{L}_{\mathrm{UB\text{-}MM}} = \frac{1}{N} \sum_{i=1}^{N} \left[c_\theta(\tau_i^{\mathrm{demo}}) - \frac{N}{M} \sum_{k=1}^{M/N} c_\theta(\tau_{i,k}^{\mathrm{samp}})\right],
    \label{eqn:moment_matching_loss_factored}
\end{equation}
where $\tau_{i,k}^{\mathrm{samp}}$ denotes the $k$-th background sample associated with the $i$-th expert demonstration. This form makes explicit that each expert demonstration is paired with $M/N$ background samples, with  $M \geq N$. 

For compactness and computational efficiency in the recursive setting, we set $M = N$, i.e., for each expert demonstration we draw one sampled trajectory from the inner control policy $q(\tau)$. Under this choice, the factor $N/M = 1$ and the inner sum in~\eqref{eqn:moment_matching_loss_factored} collapses to a single term, yielding:
\begin{equation}
    \mathcal{L}_{\mathrm{UB\text{-}MM}} = \frac{1}{N} \sum_{i=1}^{N} \left[c_\theta(\tau_i^{\mathrm{demo}}) - c_\theta(\tau_i^{\mathrm{samp}})\right].
    \label{eqn:moment_matching_loss}
\end{equation}

This upper bound has a particularly important consequence: it transforms the maximum entropy IRL objective into a moment-matching loss. This structure is equivalent to recent moment matching formulations in IRL~\citep{swamy2021moments,zeng2022maximum,zeng2025structural}, which replace the log-partition function of MaxEnt \gls{irl} with expectation-matching objectives between expert and policy distributions. Our derivation shows that moment matching losses, particularly the formulation in~\citet{swamy2021moments}, can be interpreted as an upper bound of the maximum entropy negative log-likelihood.
\medskip
\begin{remark}[Tightness of the bound]
\label{remark:tightness}
The tightness of the upper bound~\eqref{eqn:upper_bound_general} depends on the choice of cost function parameterization and sampling policy. In our experimental settings the bound remains effective, as demonstrated by the empirical results in Section~\ref{sec:exp}.
\end{remark}

\section{Recursive Deep Inverse Reinforcement Learning}
In the previous section, we derived the upper bound of the negative log-likelihood cost described in \eqref{eq:nll_mc} and showed it's equivalent to moment matching \citep{swamy2021moments}. In this section, we seek to minimize the moment matching loss of \eqref{eqn:moment_matching_loss} recursively. To do so, we re-write the EKF optimization problem using the loss function derived in \eqref{eqn:moment_matching_loss} and a regularization term.

We exploit this structure by applying a single optimization step per iteration, analogous to the EKF. At step $i$, the algorithm predicts $\hat{\theta}_{i|i-1} = \hat{\theta}_{i-1}$, then updates this prediction using the gradient of the new moment-matching term, scaled by a matrix $P_{\theta_i}$ that recursively accumulates curvature information and acts as an adaptive step size. The result is a fully online algorithm: one expert trajectory in, one parameter update out, with no storage of past data.
We first describe the optimization formulation and define all relevant quantities, then derive the recursive solution.

\subsection{Problem Formulation}


Given an expert trajectory $\dataset_\mathrm{demo} \triangleq \{\tau^{(0)},\dots,\tau^{(N-1)}\}$ we seek to determine an optimal solution $\theta^*_i$ starting from initial condition $\theta_0$ by solving the following mathematical optimization problem:

\begin{equation} 
\begin{aligned}
    \mathcal{L}_N(\Theta_N)&= \mathcal{L}_\mathrm{UB-MM} + \frac{1}{2}\sum_{i=1}^{N}\lVert \theta_i - \theta_{i-1} \rVert^{2}_{Q_{\theta}^{-1}} \\
    &= \sum_{i=1}^{N}\left[c_\theta(\tau_i^\mathrm{demo}) - c_\theta(\tau_i^\mathrm{samp})\right] + \frac{1}{2}\sum_{i=1}^{N}\lVert \theta_i - \theta_{i-1} \rVert^{2}_{Q_{\theta}^{-1}},
\end{aligned}
\label{eq:full_optimization_function_A}
\end{equation}

where $\Theta_N = \{\theta_0,\dots,\theta_N\}$ denotes the full sequence of parameter estimates. The second term penalizes large changes in $\theta$ between consecutive optimization steps, acting as a regularization controlled by $Q_\theta \in \mathbb{R}^{d_\theta \times d_\theta}$ \citep{imbiriba2022hybrid,ghanem2025learning}. The notation $\lVert x \rVert^2_{Q_\theta^{-1}} = x^\top Q_\theta^{-1} x$ denotes the energy norm with respect to $Q_\theta^{-1}$.

Before stating the recursive solution, we define all quantities used in the algorithm.
Let $\Theta_i = \{\theta_0, \ldots, \theta_i\}$ denote the full sequence of parameter estimates up to step $i$, and let $\hat{\Theta}_{i-1}$ denote the optimized sequence from the previous step. Let $\hat{\theta}_i$ denote the parameter estimate at step $i$, and let $\hat{\theta}_{i|i-1} = \hat{\theta}_{i-1}$ denote the parameter estimate before incorporating the new expert sample at step $i$. The predicted sequence is:
\begin{equation}
    \hat{\Theta}_{i|i-1}=\left[\begin{matrix} \hat{\Theta}_{i-1} \\
                        \hat{\theta}_{i|i-1} \end{matrix}\right],
\end{equation}
which extends the previous sequence by repeating the last estimate.

Let $\mathcal{L}_i(\Theta_i)$ denote the cumulative loss at step $i$ as defined in~\eqref{eq:full_optimization_function_A}. We define the following gradient and Hessian quantities of the cost function, evaluated at $\hat{\theta}_{i-1}$:
\begin{align}
C_{\tau_\mathrm{demo}}(i) &= \frac{\partial c_{\theta}(\tau_{i}^\mathrm{demo})}{\partial \hat{\theta}_{i-1}}, &
C_{\tau_\mathrm{samp}}(i) &= \frac{\partial c_{\theta}(\tau_{i}^\mathrm{samp})}{\partial \hat{\theta}_{i-1}}, \label{eq:gradient_defs} \\
C_{\tau_\mathrm{demo}}^2(i) &= \frac{\partial^2 c_{\theta}(\tau_{i}^\mathrm{demo})}{\partial^2 \hat{\theta}_{i-1}}, &
C_{\tau_\mathrm{samp}}^2(i) &= \frac{\partial^2 c_{\theta}(\tau_{i}^\mathrm{samp})}{\partial^2 \hat{\theta}_{i-1}}. \label{eq:hessian_defs}
\end{align}
Additionally, we define:
\begin{itemize}
    \item $P_{\theta_i} \in \mathbb{R}^{d_\theta \times d_\theta}$: the lower-right block of the full inverse Hessian $\left(\nabla^2 \mathcal{L}_i(\hat{\Theta}_{i|i-1})\right)^{-1}$, representing accumulated second-order curvature information and acting as an adaptive step-size matrix; initialized as $P_{\theta_0} = 10^{-2} I$.
    \item $Q_\theta \in \mathbb{R}^{d_\theta \times d_\theta}$: the regularization matrix, initialized as $Q_\theta = 10^{-4} I$, which prevents $P_{\theta_i}$ from collapsing to zero and keeps the algorithm responsive to new data.
    
\end{itemize}

\subsection{Derivation of the Recursive Solution}

Inspired by the recursive second-order optimization framework described in \citet{humpherys2012fresh}, we seek to determine the optimal solution $\Theta_N^*=\{\theta^*_0,\dots,\theta^*_N\}$ using the second-order Newton method sequentially, which recursively finds $\Theta_N^*$ given $\Theta_{N-1}^*$.

At each step $i$, RDIRL minimizes the cumulative loss $\mathcal{L}_i(\Theta_i)$ over the full sequence $\Theta_i = \{\theta_0, \ldots, \theta_i\}$ by performing one Newton step. Because the loss has the recursive structure of~\eqref{eq:full_optimization_function_A}, this Newton step can be computed using only $\hat{\Theta}_{i-1}$ and the new pair $(\tau^{\mathrm{demo}}_i, \tau^{\mathrm{samp}}_i)$, without revisiting past data. Of the updated sequence $\hat{\Theta}_i$, only the last block $\hat{\theta}_i$ is retained as the new parameter estimate; earlier blocks remain unchanged from $\hat{\Theta}_{i-1}$. The matrix $P_{\theta_i}$ accumulates second-order curvature information and acts as an adaptive step size.

We start by breaking the optimization function~\eqref{eq:full_optimization_function_A} as follows:
\begin{equation} 
\begin{aligned}
\mathcal{L}_{i}(\Theta_i)=\mathcal{L}_{i-1}(\Theta_{i-1}) + c_\theta(\tau_i^\mathrm{demo}) - c_\theta(\tau_i^\mathrm{samp}) + \frac{1}{2}\lVert \theta_i -\theta_{i-1} \rVert^{2}_{Q_{\theta}^{-1}}.
\end{aligned}
\label{eq:divided_optimization_function_A}
\end{equation}

Next, we further divide~\eqref{eq:divided_optimization_function_A} into a prediction step and an update step:
\begin{equation} 
\begin{aligned}
\mathcal{L}_{i}(\Theta_i)&= \mathcal{L}_{i|i-1}(\Theta_i)+ c_\theta(\tau_i^\mathrm{demo}) - c_\theta(\tau_i^\mathrm{samp}), 
\end{aligned}
\label{eq:divided_optimization_function_B}
\end{equation}
where
\begin{equation} 
\mathcal{L}_{i|i-1}(\Theta_i)= \mathcal{L}_{i-1}(\Theta_{i-1}) + \frac{1}{2}\lVert \theta_i - \theta_{i-1} \rVert^{2}_{Q_{\theta}^{-1}}.  
\label{eq:prediction_optimization_function_theta}
\end{equation}

Our optimization approach consists of first minimizing~\eqref{eq:prediction_optimization_function_theta} to obtain the predictor $\hat{\Theta}_{i|i-1}$, then minimizing~\eqref{eq:divided_optimization_function_B} given this predictor. We minimize~\eqref{eq:prediction_optimization_function_theta} with respect to $\Theta_i$ by finding $\Theta_i$ that drives the gradient to zero. Taking the gradient of~\eqref{eq:prediction_optimization_function_theta} with respect to $\Theta_i$, we obtain:
\begin{equation} 
\begin{aligned}
\nabla \mathcal{L}_{i|i-1}(\Theta_i)=\left[\begin{matrix} 
    \nabla \mathcal{L}_{i-1}(\Theta_i) - L_\theta^{T}Q_{\theta}^{-1}[\theta_{i}-\theta_{i-1}] \\
    Q_{\theta}^{-1}[\theta_{i}-\theta_{i-1}]
    \end{matrix}\right]
\label{eq:gradient_prediction_theta}
\end{aligned}
\end{equation}
with $L_\theta=[0_{d_\theta\times d_\theta},\dots,0_{d_\theta\times d_\theta},I_{d_\theta\times d_\theta}]$ where $L_\theta \in \mathbb{R}^{d_\theta \times ((i-1)\times d_\theta)}$.

Setting $\nabla \mathcal{L}_{i|i-1}(\Theta_i) = 0$, the minimizer $\hat{\Theta}_{i|i-1}$ can be decomposed as:
\begin{equation}
    \begin{aligned}
     \hat{\Theta}_{i|i-1}=\left[\begin{matrix} \hat{\Theta}_{i-1} \\
                        \hat{\theta}_{i-1} \end{matrix}\right].
    \end{aligned}
    \label{eq:theta_hat}
\end{equation}

Given~\eqref{eq:theta_hat}, we proceed to minimize~\eqref{eq:divided_optimization_function_B} using the second-order Newton update. The gradient of~\eqref{eq:divided_optimization_function_B} at $\Theta_i = \hat{\Theta}_{i|i-1}$ is:
\begin{equation} 
\nabla \mathcal{L}_{i}(\Theta_i)=\left[\begin{matrix} 
   \mathbf{0}_{(i-1) \cdot d_\theta} \\
    C_{\tau_\mathrm{demo}}(i) -  C_{\tau_\mathrm{samp}}(i)
    \end{matrix}\right]
\label{eq:gradient_theta_2}
\end{equation}
and the Hessian of~\eqref{eq:divided_optimization_function_B} is:
\begin{equation}
\begin{aligned}
\nabla^2 \mathcal{L}_{i}(\Theta_i)= \left[\begin{matrix} \nabla^2 \mathcal{L}_{i-1}(\Theta_{i-1}) + Q_{\theta}^{-1}  & -L_{\theta}^{T}Q_{\theta}^{-1} \\ 
-Q_{\theta}^{-1}L_{\theta} & Q_{\theta}^{-1} + C_{\tau_\mathrm{demo}}^2(i) -  C_{\tau_\mathrm{samp}}^2(i)\end{matrix}\right].
\end{aligned}
\label{eq:hessian_full}
\end{equation}

Using the Newton second-order method, we update our estimate of $\Theta_{i}$ given $\hat{\Theta}_{i|i-1}$ as follows:
\begin{equation}
\hat{\Theta}_i=\hat{\Theta}_{i|i-1} -\left(\nabla^2 \mathcal{L}_{i}(\hat{\Theta}_{i|i-1})\right)^{-1}\nabla \mathcal{L}_{i}(\hat{\Theta}_{i|i-1}).
\label{eq:newton_equation_theta}
\end{equation} 

The resulting optimal variable $\hat{\theta}_i \in \hat{\Theta}_i$ is obtained by extracting the last $d_\theta$-dimensional block of~\eqref{eq:newton_equation_theta}, which leads to the following theorem.  

\begin{theorem}\label{thm:rdirl}
Given $\hat{\theta}_{i-1} \in \hat{\Theta}_{i-1}$ and known $P_{\theta_{i-1}} \in \mathbb{R}^{d_\theta \times d_\theta}$, the recursive equations for computing $\hat{\theta}_i$ that minimizes~\eqref{eq:divided_optimization_function_B} are given by the following:
\begin{equation}
    \hat{\theta}_i = \hat{\theta}_{i|i-1} - P_{\theta_i}(C_{\tau_\mathrm{demo}}(i) - C_{\tau_\mathrm{samp}}(i))
\label{eq:theta_update}
\end{equation}
where $\hat{\theta}_{i|i-1} = \hat{\theta}_{i-1}$ is the predicted parameter. $P_{\theta_i}$ being the lower right block of $\left(\nabla^2 \mathcal{L}_i(\hat{\Theta}_{i|i-1})\right)^{-1}$ recursively calculated as:
\begin{equation}
    P_{\theta_i} = \left[(P_{\theta_{i-1}} + Q_\theta)^{-1} + \left(C_{\tau_\mathrm{demo}}^2(t_i) - C_{\tau_\mathrm{samp}}^2(t_i)\right)\right]^{-1}
\label{eq:P_update}
\end{equation}
\end{theorem}

\begin{proof}
By the Schur complement formula (block matrix inversion lemma) applied to the Hessian~\eqref{eq:hessian_full}, the lower-right block of $\left(\nabla^2 \mathcal{L}_i(\hat{\Theta}_{i|i-1})\right)^{-1}$ is:
\begin{equation}
    P_{\theta_i} = \left[ \left(P_{\theta_{i-1}} + Q_\theta\right)^{-1} + C^2_{\tau_\mathrm{demo}}(i) - C^2_{\tau_\mathrm{samp}}(i) \right]^{-1},
\end{equation}
where $P_{\theta_{i-1}} = \left(\nabla^2 \mathcal{L}_{i-1}\right)^{-1}_{\theta\theta}$ is the lower-right block of the previous inverse Hessian, and the term $(P_{\theta_{i-1}} + Q_\theta)^{-1}$ arises from the regularization term in~\eqref{eq:prediction_optimization_function_theta}. This derivation applies Lemma~B.3 of \citet{humpherys2012fresh}. Substituting this into the last $d_\theta$ rows of the Newton update~\eqref{eq:newton_equation_theta}, together with the gradient~\eqref{eq:gradient_theta_2}, yields~\eqref{eq:theta_update}.
\end{proof}

As a consequence of Theorem~\ref{thm:rdirl}, $\hat{\theta}_i$ is computed according to~\eqref{eq:theta_update} using only $\hat{\theta}_{i-1}$, the new trajectories $(\tau_i^{\mathrm{demo}}, \tau_i^{\mathrm{samp}})$, and the accumulated inverse Hessian $P_{\theta_{i-1}}$. The entire training procedure is detailed in Algorithm~\ref{alg:cap}.

\subsection{Algorithm Description}

Algorithm~\ref{alg:cap} maintains a cost function $c_\theta(\tau)$ parameterized by $\theta$, which maps trajectories $\tau$ to scalar costs. The goal is to iteratively update $\theta$ such that trajectories generated from the current policy $q(\tau)$ match the expert demonstrations.

At each outer iteration (episode), we initialize the sampling policy $q(\tau)$, which can be any stochastic policy optimized with methods like PPO or MPPI. We also initialize the inverse Hessian $P_{\theta_{0}}$ and the regularization matrix $Q_\theta$. The matrix $P_{\theta_0}$ represents the initial step-size scaling for the Newton updates, and $Q_\theta$ controls the regularization strength, preventing $P_{\theta_i}$ from collapsing to zero across iterations.

For each inner iteration, as soon as the algorithm observes one expert demonstration $\tau_i^{\mathrm{demo}}$, it samples a trajectory $\tau_i^{\mathrm{samp}}$ from $q(\tau)$. The gradients $C_{\tau_\mathrm{demo}}(i)$ and $C_{\tau_\mathrm{samp}}(i)$, as well as the Hessians $C_{\tau_\mathrm{demo}}^2(i)$ and $C_{\tau_\mathrm{samp}}^2(i)$, are then computed. The parameter $\theta$ is updated via~\eqref{eq:theta_update} and the inverse Hessian via~\eqref{eq:P_update}. After updating $\theta$, the sampling policy $q(\tau)$ is improved using any standard policy optimization method (e.g., PPO, MPPI), guided by the updated cost function $c_\theta$. This process continues over $K$ episodes, gradually aligning the agent's behavior with that of the expert.

\begin{algorithm}[t]
\caption{Recursive Deep Inverse Reinforcement Learning}\label{alg:cap}

\textbf{Input:} Cost function $c_\theta$ with randomly initialized parameters $\theta_0$, 
regularization matrix $Q_\theta$, initial step-size matrix $P_{\theta_0}$, 
number of episodes $K$, number of expert samples per episode $N$ \\
\While{$\mathrm{episodes} < K$}{
    Initialize inner policy $q(\tau)$\;
    
    \For{$i = 1, 2, \ldots, N$}{
        Observe one expert sample $\tau_i^{\mathrm{demo}}$\;
        Sample one trajectory $\tau_i^{\mathrm{samp}}$ from $q(\tau)$\;
        Evaluate the gradients $C_{\tau_\mathrm{demo}}(i)$ and $C_{\tau_\mathrm{samp}}(i)$\;
        Evaluate the Hessians $C_{\tau_\mathrm{demo}}^2(i)$ and $C_{\tau_\mathrm{samp}}^2(i)$\;
        $P_{\theta_i} \gets \left[(P_{\theta_{i-1}} + Q_\theta)^{-1} + C_{\tau_\mathrm{demo}}^2(i) - C_{\tau_\mathrm{samp}}^2(i)\right]^{-1}$\;
        $\hat{\theta}_i \gets \hat{\theta}_{i-1} - P_{\theta_i}\left(C_{\tau_\mathrm{demo}}(i) - C_{\tau_\mathrm{samp}}(i)\right)$\;

        Update $q(\tau)$ w.r.t.\ $c_\theta$ using any policy optimization method\;
    }
    $\mathrm{episodes} \gets \mathrm{episodes} + 1$\;
}
\end{algorithm}

\section{Experiments}
\label{sec:exp}


We evaluate the proposed \gls{rdirl} algorithm in continuous control benchmarks from OpenAI Gym \citep{brockman2016openai} and MuJoCo \citep{todorov2012mujoco}, as well as in an adversarial cognitive radar scenario ~\citep{potter2024continuously,haykin2006cognitive}. We compare its performance against state-of-the-art inverse reinforcement learning and imitation learning methods, including \gls{gail} \citep{ho2016generative}, \gls{gcl} \citep{finn2016guided}, \gls{airl} \citep{fu2017learning}, SQIL \citep{reddy2019sqil}, and \gls{mlirl} \citep{zeng2022maximum}, a moment-matching variant of \gls{irl}. Experiments are conducted in two regimes: batch mode (section \ref{sec:exp}), where competing methods are trained in their standard setting with full trajectory batches, and streaming mode, where updates occur sample by sample (Appendix~\ref{app:additional_experiments}).

Unlike reinforcement learning methods such as SAC \citep{haarnoja2018soft} or PPO \citep{schulman2017proximal}, which require large trajectory batches to converge and thus fail in streaming or real-time settings, our approach leverages \gls{mppi} \citep{williams2016aggressive} as the inner control policy. Since \gls{mppi} updates its actions at every time step, it is naturally suited for online \gls{irl}. In preliminary experiments, \gls{mppi} also provided stable performance and fast convergence unlike traditional RL policies when integrated into the \gls{rdirl} framework. Furthermore, preliminary experiments showed that competing IRL methods paired with their original RL inner policies failed to converge in streaming mode too. For consistency and fairness, we therefore adapt all competing methods to use \gls{mppi} as the inner policy in both batch and streaming comparisons.

Our results show that \gls{rdirl} consistently outperforms all benchmarked methods in recovering reward functions. Policies trained with rewards learned by \gls{rdirl} achieve optimal or near-optimal behavior significantly faster than competing approaches. Crucially, unlike existing methods which require large batches of expert trajectories and environment rollouts to converge, \gls{rdirl} leverages online adaptation. This enables efficient learning from streaming demonstrations, making it particularly well-suited for adversarial and time-limited scenarios.

\begin{table}[ht]
\centering
\caption{Comparison of normalized averaged reward values across all episodes for different Gym environments and methods.}
\label{tab:scores}
\setlength{\tabcolsep}{3.6pt}
\begin{tabular}{p{3.5cm}|p{2.2cm}|p{3.0cm}|p{2.6cm}|p{2.5cm}}
\toprule
Methods & CartPole & MountainCar & HalfCheetah-v4 & Hopper \\ 
\midrule
SQIL     & $0.947\pm0.088$      & $-0.001 \pm 4.79\times 10^{-5}$ & $-1.56 \pm 0.89$    & $0.799\pm0.15$    \\
GAIL  & $0.934\pm0.058$      & $0.236\pm0.203$                         & $-0.521\pm1.15$     & $0.714\pm0.08$    \\
GCL     & $0.92\pm0.09$        & $0.247\pm0.19$                          & $-0.226\pm1.27$     & $0.69\pm0.075$    \\
AIRL    & $0.953\pm0.069$      & $0.233\pm0.204$                         & $-0.54\pm1.11$      & $0.709\pm0.084$   \\
ML-IRL & $0.938\pm0.093$      & $0.253\pm0.19$                          & $-0.32\pm1.12$      & $0.648\pm0.06$    \\
\midrule
RDIRL (ours)                  & $\bm{0.993\pm0.013}$ & $\bm{0.68\pm0.32}$                      & $\bm{0.496\pm0.59}$ & $\bm{0.803\pm0.11}$ \\
\bottomrule
\end{tabular}
\end{table}

\begin{figure*}[h]
\centering

\includegraphics[width=1.0\textwidth]{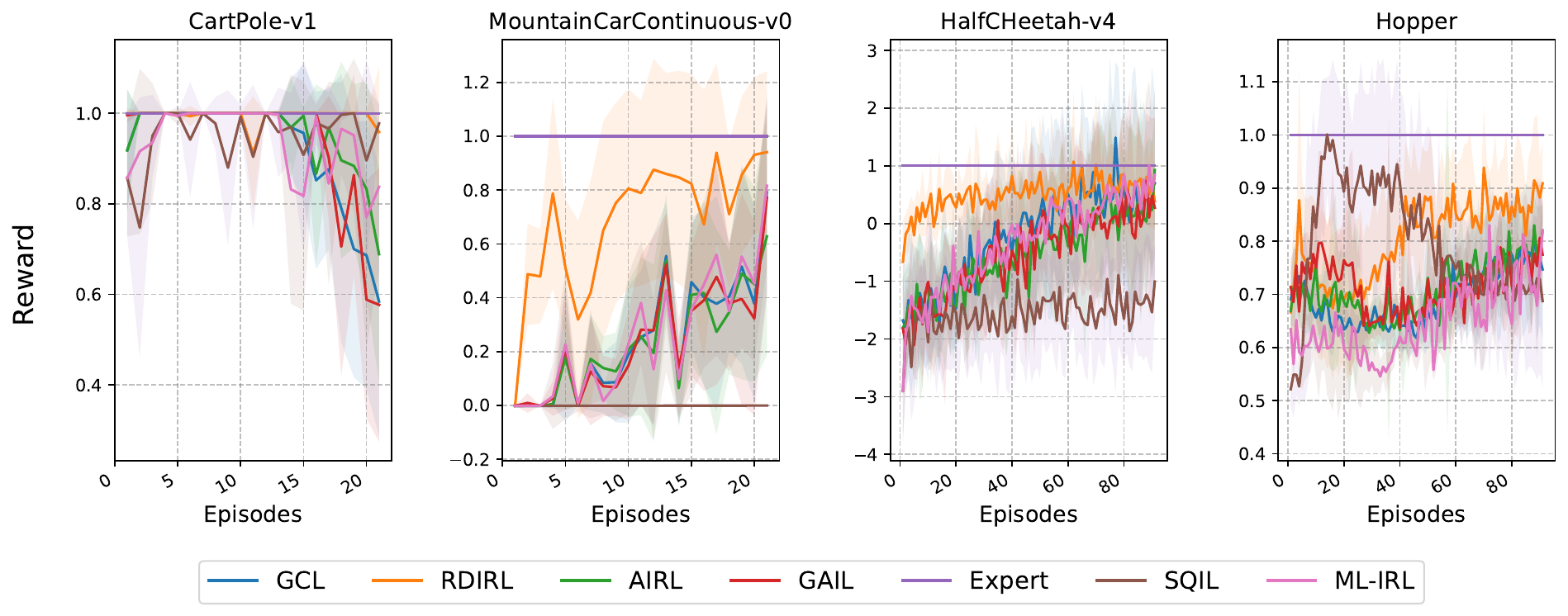}
\caption{Learning curves for RDIRL and other methods.}
\label{fig:results_continous}
\end{figure*}
\subsection{Continuous control}

To assess the performance of our proposed approach \gls{rdirl}, we conduct inverse reinforcement learning (IRL) experiments on the CartPole and Mountain Car environments from OpenAI Gym ~\citep{brockman2016openai} and HalfCheetah-v4, Hopper, and Walker2d from MuJoCo \citep{todorov2012mujoco}, all solved using model-free reinforcement learning. Each task has a predefined true reward function provided by OpenAI Gym.

We first generate expert demonstrations for these tasks by training a PPO reinforcement learning agent ~\citep{schulman2017proximal} to maximize the true reward function. Each expert demonstration consists of a state trajectory of size $N$ steps specified in Table \ref{tab:table of parameters} in \ref{app:training} for each task, which is then used as the sole expert trajectory for each IRL algorithm. Note that we only use the expert's state trajectory and not the expert's actions, since we do not have access to the expert's control policy.

Next, we execute \gls{rdirl} to learn the reward function and train competing \gls{irl} algorithms using the expert trajectory over multiple episodes in batch mode, where each episode consists of an expert trajectory. This process is repeated for 12 Monte Carlo runs with different seeds. In all experiments, we use \gls{mppi} as the internal control policy $q(\tau)$ to maximize the learned reward function, $-c_{\theta}$. A detailed experiment description and parameter values of \gls{mppi} and \gls{irl} algorithms is described in Appendix \ref{app:training} 

We plot the mean of the normalized cumulative reward values across all episodes of trajectories $\tau^\mathrm{samp}$  sampled from the inner control policy $q(\tau)$ in Figure \ref{fig:results_continous}.the averaged reward values are normalized with respect to the expert reward. In the case of \gls{rdirl}, $\tau^\mathrm{samp}$ used to calculate the reward function in Figure \ref{fig:results_continous} are generated online during training according to Algorithm \ref{alg:cap}. For the rest of the methods, $\tau^\mathrm{samp}$ are generated offline after each offline training episode is completed.

All methods use the same neural network architecture to parameterize the reward function. Networks are randomly initialized at the start of each experiment, and all experiments are run on Nvidia-H200 GPU Cluster with 1 GPU per job(seed).

\begin{figure}[t]
  \centering
  \includegraphics[width=0.48\textwidth]{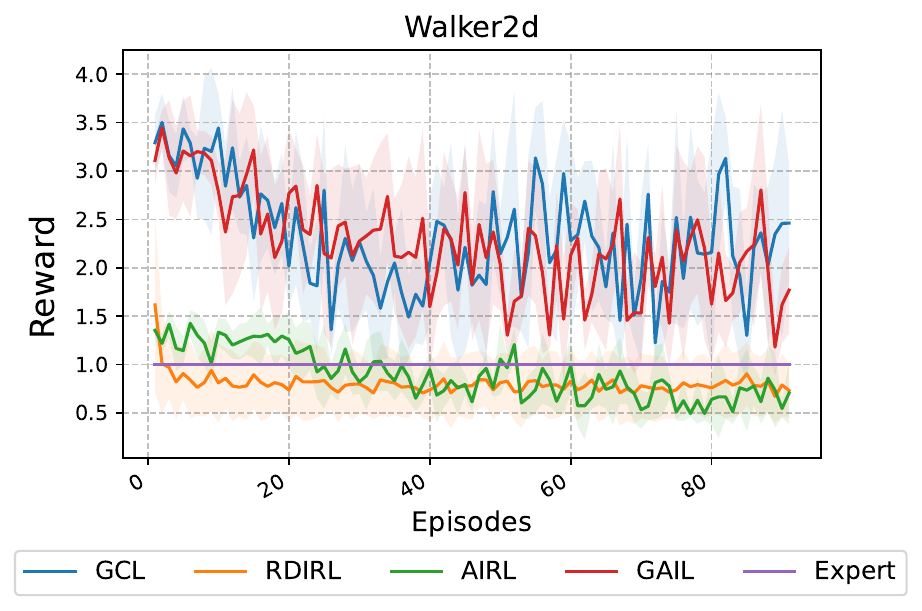}
  \caption{Learning curves for for Walker2d.}\label{fig:Walker2d}
\end{figure}

Our proposed method, \gls{rdirl}, successfully learns reward functions across all benchmark environments and consistently outperforms competing methods. In CartPole and MountainCar, it quickly recovers the expert reward even converging in one episode in CartPole, while in HalfCheetah and Hopper it achieves faster convergence and higher reward quality than baselines, many of which require far more episodes to converge or fail to converge. Learning curves in Figures~\ref{fig:results_continous} and~\ref{fig:Walker2d} illustrate these improvements, with Walker2d results consistent with Figure 3 in \citet{reddy2019sqil}, where rewards closer to the expert indicate better performance. Furthermore, experimental results in streaming settings with detailed descriptions are provided in Appendix \ref{app:additional_experiments}.

Table~\ref{tab:scores} further shows that \gls{rdirl} achieves the highest normalized rewards in most tasks. This consistent outperformance stems from its recursive structure and adaptive uncertainty-aware updates, which improve sample efficiency and stability. Unlike traditional IRL, our method requires no fixed learning rate, as $P_{\theta}$ is updated at each step and acts as an adaptive rate.

\subsection{Cognitive radar}
To evaluate whether our method can learn cost functions of adversarial agents, we perform inverse reinforcement learning experiments on a cognitive radar task following the setup of \citet{potter2024continuously}. The task involves a radar chasing a moving target in 3D space.

\textbf{Expert (Radar) Dynamics and Reward.}
The radar follows the second-order unicycle model of \citet{potter2024continuously}, with state $\chi^R_k = [x, y, z, \theta, v, \omega]$ representing 3D Cartesian position, heading angle $\theta$, heading velocity $v$, and angular velocity $\omega$. The discrete-time kinematic model is:
\begin{equation}
    \chi^R_{k+1} = \chi^R_k + G_k(\chi^R_k, u_k),
\end{equation}
where
\begin{equation}
    G_k(\chi^R_k, u_k) = \begin{bmatrix} v_k \cos(\theta_k) \\ v_k \sin(\theta_k) \\ 0 \\ \omega_k \\ u_a \\ u_{\dot{\omega}} \end{bmatrix} \Delta t,
\end{equation}
with control inputs $u = [u_a, u_{\dot{\omega}}]^\top$ (linear and angular acceleration) subject to the constraints $\underline{u}_a \leq u_{a_k} \leq \bar{u}_a$ and $\underline{u}_{\dot{\omega}} \leq u_{\dot{\omega}_k} \leq \bar{u}_{\dot{\omega}}$ for all $k$, and $\Delta t = 0.1$~s is the control timestep. We denote the 3D position components of the radar and target states as $\chi^R_{xyz} = [x, y, z]^\top$ and $\chi^T_{xyz} = [x, y, z]^\top$, respectively.

The radar's true reward function is the log-determinant of the Standard \gls{fim} \citep{potter2024continuously} for target localization:
\begin{equation}
    r = \log \det J(\chi^T_{xyz};\, \chi^{R}_{xyz}),
\end{equation}
where the \gls{fim} for a single radar-target pair is given by \citet{potter2024continuously}:
\begin{equation}\label{eq:fim_radar}
    J(\chi^T_{xyz}; \chi^R_{xyz}) = (\chi^T_{xyz} - \chi^R_{xyz})(\chi^T_{xyz} - \chi^R_{xyz})^\top \left( \frac{4}{\Gamma \|\chi^R_{xyz} - \chi^T_{xyz}\|^6} + \frac{8}{\|\chi^R_{xyz} - \chi^T_{xyz}\|^4} \right),
\end{equation}
with $\Gamma = \frac{\sigma_a^2 \pi L}{2 P_t \Lambda_t \Lambda_r \Xi}$ determined by the radar signal parameters: transmit power $P_t = 1000$~W, transmit and receive gains $\Lambda_t = \Lambda_r = 200$, radar cross section $\Xi = 1$~m$^2$, loss factor $L=1$, carrier frequency $f_c = 10^8$~Hz, and noise power $\sigma_a^2$ set such that $\mathrm{SNR} = -20$~dB at range $R = 500$~m. The radar expert policy is an \gls{mppi} controller that maximizes this \gls{fim} reward with horizon $H=10$, number of sampled trajectories $= 25$, and temperature $= 10^{-2}$.

\textbf{Target Dynamics and Learned Reward.}
The target follows a constant velocity motion model with acceleration noise \citep{baisa2020derivation,potter2024continuously}, with state $\chi^T_k = [x, y, z, \dot{x}, \dot{y}, \dot{z}]$ and transition model:
\begin{equation}
    \chi^T_{k+1} = A_{\mathrm{single}}\, \chi^T_k + \epsilon_w, \quad
    A_{\mathrm{single}} = \begin{bmatrix} \mathbf{I}_{3\times3} & \Delta t\, \mathbf{I}_{3\times3} \\ \mathbf{0}_{3\times3} & \mathbf{I}_{3\times3} \end{bmatrix},
\end{equation}
where $\epsilon_w \sim \mathcal{N}(0, W_{\mathrm{single}})$ with $W_{\mathrm{single}} = W_{\Delta t}\, \Sigma_w\, W_{\Delta t}^\top$, $W_{\Delta t} = \begin{bmatrix} \frac{1}{2}\Delta t^2\, \mathbf{I}_{3\times3} \\ \Delta t\, \mathbf{I}_{3\times3} \end{bmatrix}$, $\Sigma_w = \sigma_W^2\, \mathbf{I}_{3\times3}$, and $\sigma_W = \sqrt{10}$.

The target's goal is to learn the radar's \gls{fim} reward function online using \gls{rdirl}. The learned cost function $c_\theta$ is parameterized by a one-hidden-layer MLP with 128 units and ReLU activation, taking as input the relative displacement $\Delta p = \chi^R_{xyz} - \chi^T_{xyz} \in \mathbb{R}^3$ between the radar and target 3D positions. The output of the network is passed through a ReLU activation to ensure non-negative cost values.

To achieve this goal, we execute Algorithm~\ref{alg:cap} where the learned cost function approximating the radar's reward is updated online.  The inner control policy $q(\tau)$ is an \gls{mppi} that maximizes the learned reward function, $-c_{\theta}$. Additional environment and \gls{irl} method's parameters are described in Table~\ref{tab:table of parameters}.
\begin{table}[ht]
\centering
\caption{Comparison of mean FIM reward values for the Cognitive Radar example obtained by the different IRL methods.}
\label{tab:scores_radar}
\setlength{\tabcolsep}{6.6pt}
\begin{tabular}{p{3.5cm}|c}
\toprule
Methods & Mean Cumulative Reward \\
\midrule
GAIL   & $153.05$ \\
GCL    & $423.49$ \\
AIRL   & $196.53$ \\
\midrule
RDIRL (ours) & $\bm{924.78}$ \\
\bottomrule
\end{tabular}
\end{table}
Furthermore, we compare \gls{rdirl} against \gls{gail}, \gls{airl}, and \gls{gcl}. To implement these methods, we generate expert trajectories for multiple episodes, where the expert policy is an \gls{mppi} that maximizes the radar's \gls{fim}. The inner control policy $q(\tau)$ in all of these baselines is an \gls{mppi}, with parameters specified in Table~\ref{tab:table of parameters}. We repeat this process for 5 Monte Carlo runs using different seeds.
\begin{figure}[t]
  \centering
  \includegraphics[width=0.48\textwidth]{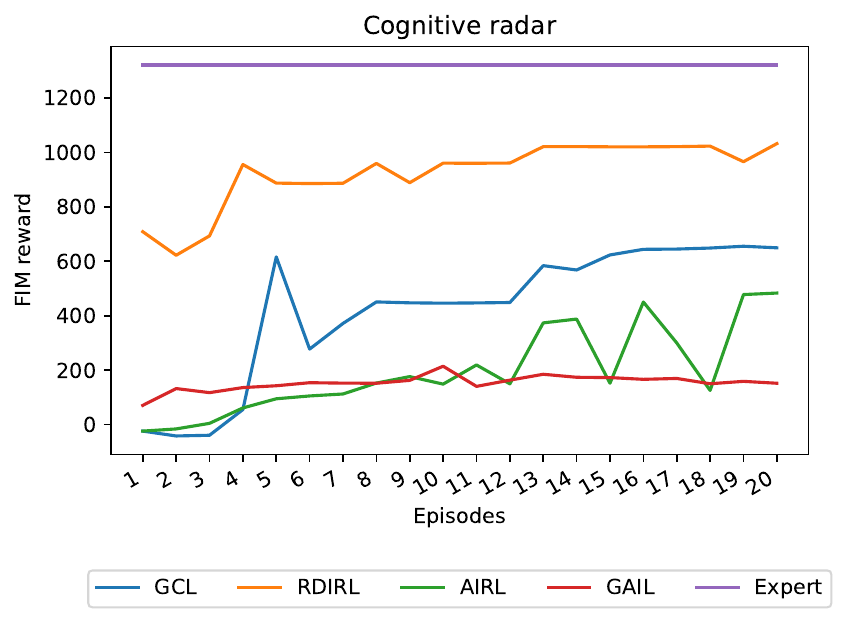}
  \caption{Learning curves for RDIRL and other methods.}
  \label{fig:results_radar}
\end{figure}
To test if the target successfully learned the radar's reward function, we execute the radar using an \gls{mppi} policy that maximizes the learned reward function $-c_\theta$ instead of the true \gls{fim}. We  plot the cumulative true \gls{fim} values resulting from the trajectories $\tau^{\mathrm{samp}}$ sampled from the inner control policy $q(\tau)$ in Figure~\ref{fig:results_radar}. We compare \gls{rdirl}'s performance in learning the radar's reward function against \gls{gail}, \gls{gcl}, and \gls{airl}. In the case of \gls{rdirl}, $\tau^\mathrm{samp}$ used to evaluate  the learned reward function in Figure~\ref{fig:results_radar} are generated online during training according to Algorithm~\ref{alg:cap}. For the rest of the methods, $\tau^\mathrm{samp}$ are generated offline after each offline training episode is completed. In all algorithms, we used the same neural network architecture described above to parameterize the learned reward function: one hidden layer of 128 units, with a ReLU activation function. All networks were always initialized randomly at the start of each experiment and all experiments are run on an Intel Core i7 CPU.

Results in Figure~\ref{fig:results_radar} show that \gls{rdirl} successfully learns the radar's \gls{fim} with a much faster convergence rate than the benchmark methods. The mean cumulative reward values across all episodes for each method are summarized in Table~\ref{tab:scores_radar}. As shown, \gls{rdirl} outperforms all other methods in terms of the mean cumulative reward, significantly outperforming the benchmark methods (i.e., \gls{airl}, \gls{gcl}, and \gls{gail}).

\vspace{1em}
\section{Conclusions}
We presented \gls{rdirl} within the \gls{irl} framework that generalizes recent advances in maximum entropy deep \gls{irl} to online settings. We first established the equivalence between upper bound loss function in \eqref{eqn:moment_matching_loss} of the negative log likelihood in \eqref{eq:nll_mc} to moment matching loss of \citep{swamy2021moments}. Second, we leveraged sequential second-order Newton optimization to derive an online \gls{irl} algorithm by minimizing the moment matching loss function  of \eqref{eqn:moment_matching_loss} recursively and therefore established key theoretical properties of maximum entropy online deep \gls{irl}

\gls{rdirl} can learn rewards and cost functions online and greatly outperforms both prior imitation learning and \gls{irl} algorithms in terms of steps and samples required to converge. It generally reproduces the batch method’s accuracy but in significantly less steps.

The recursive updates in Algorithm~\ref{alg:cap} require computing the full Hessian of the cost function at each step, which scales quadratically in the number of parameters $d_\theta$. In our experiments this is tractable because the networks are small (two layers of 16 units for continuous control, one layer of 128 units for cognitive radar). For larger networks, well-established approximations such as Gauss-Newton, block-diagonal Hessian approximations~\citep{zhang2018block}, or Kronecker-factored preconditioners such as Shampoo~\citep{gupta2018shampoo,anil2020scalable} can serve as drop-in replacements for the exact Hessian in~\eqref{eq:P_update}, preserving the recursive structure while reducing computational cost. We plan to explore these extensions in future work.

\section*{Acknowledgments}
This work was supported in part by the Army Research Laboratory (ARL) under cooperative agreement W911NF2320014, and by the National Institutes of Health (NIH) under grants 7R01DK133605 and R01LM014191.

\bibliography{tmlr2026_conference}
\bibliographystyle{tmlr}

\appendix
\section{Appendix}

\section{Experiment details}

In this section, we list down the implementation details of \gls{rdirl} and the baselines. The code is
included in the supplementary material. We also report the hyperparameters used in the experiments,
the detailed network architectures, training procedures and evaluation procedures used for our
experiments.

\subsection{Training}
\label{app:training}
In all our experiments, we use \gls{mppi}\citep{williams2016aggressive}  as inner policy $q(\tau)$ in our baseline methods. \gls{mppi}is a probabilistic model predictive control policy that estimates an optimal action distribution that minimizes an agent's objective cost function. To do so, \gls{mppi} samples a number of trajectories and weighs these trajectories depending on how well they minimize the cost function, then updates the mean of its action distribution $q(\tau)$ accordingly. Since \gls{mppi} in an online policy , i.e it updates itself every time step, it makes it a natural choice of inner policy for online \gls{irl} problems, as we noticed in our preliminary experiments that it is much more stable and has faster convergence that traditional RL methods when implemented inside \gls{rdirl}.

The implementation of the baselines (GCL, AIRL,SQIL,ML-IRL and GAIL) are adapted from available public 
repository \citep{imitation2021}.Furthermore, we adapt all the baselines to use  \gls{mppi} as inner policy alongside our proposed approach. Since the inner policy  is not SAC anymore like it was in the original baselines repositories, we tune the parameters of all the adapted baselines using grid search to produce best possible performance. The resulting parameters were used directly in \gls{rdirl}.
We list the hyper-parameters of all the baselines used in different environments in Table \ref{tab:table of parameters}. These hyper-parameters were selected via
grid search.

\begin{table}[ht]
\centering
\small
\caption{List of parameters used in each environment.}
\label{tab:table of parameters}
\setlength{\tabcolsep}{4pt}
\begin{tabular}{p{2.4cm}|c|c|c|c|c|c|c}
\toprule
Environment & Learning rate & Batch size & Reward updates & $N_{\text{steps}}$ & Temperature & Horizon & Trajectories \\
\midrule
CartPole-v1      & \num{1e-4} & 150 & 15 & 150 & \num{1e-3} & 50  & 2000 \\
MountainCar-v0   & \num{1e-4} & 200 & 15 & 200 & \num{1e-2} & 85  & 3500 \\
HalfCheetah-v4   & \num{1e-4} & 200 & 15 & 200 & \num{1e-2} & 50  & 500  \\
Walker2d         & \num{1e-4} & 200 & 15 & 200 & \num{1e-2} & 50  & 500  \\
Hopper           & \num{1e-4} & 200 & 15 & 200 & \num{1e-2} & 50  & 500  \\
Cognitive Radar  & \num{1e-4} & 200 & 10 & 200 & \num{1e-2} & 10  & 25   \\
\bottomrule
\end{tabular}
\end{table}

In all our experiments, we do multiple passes of parameter updates at the end of each episode using the Adam optimizer for all the baselines for best performance, except in our proposed approach \gls{rdirl}, since it is online. The number of passes is listed in the reward function update column of \ref{tab:table of parameters}. The number of steps executed in each episode in listed in Nsteps column. Temperature,horizon and number of sampled trajectories are \gls{mppi} parameters.

PPO \citep{schulman2017proximal}  is used as the base MaxEnt RL
algorithm for the expert policy. Adam is used as the optimizer. 

In our proposed \gls{rdirl}, we use the same parameters of \ref{tab:table of parameters}. Additionally, we use $P_{\theta_0} = 10^{-2}I$ and $Q_\theta = 10^{-4}I$ where $I$ is the identity matrix.

\subsection{Reward Function and Discriminator Network Architectures}
We use the same neural network architecture to parameterize the cost function/reward 
function/discriminator for all methods. For continuous control task with raw state input, i.e. Cartpole,MountainCar,
 and the MuJoCo tasks, we use two-layer of MLP with ReLU activation function to parameterized the cost function/discriminator with a hidden size of (16,16). Networks are randomly initialized at the start of each experiment, and all experiments are run on Nvidia-H200 GPU Cluster with 1 GPU per job(seed), with runtimes ranging from 30s/episode for CartPole and 2mins/episode for Walker2d on all benchmarked and competing \gls{irl} methods.

\subsection{Online Adaptation of competing methods}
\label{app:additional_experiments}
In this section, we compare our proposed approach, \gls{rdirl}, with online-adapted versions of \gls{gail}, \gls{airl},\gls{mlirl}, and \gls{gcl}. The online adaptation involves training each competing method using one expert demonstration at a time. Specifically, the loss function of each method is computed using a single observed expert sample at each time step, followed by an immediate update of the reward function neural network parameters. This process is repeated across the full episode of Nsteps.

\begin{figure}[H]
\centering

\includegraphics[width=1.0\textwidth]{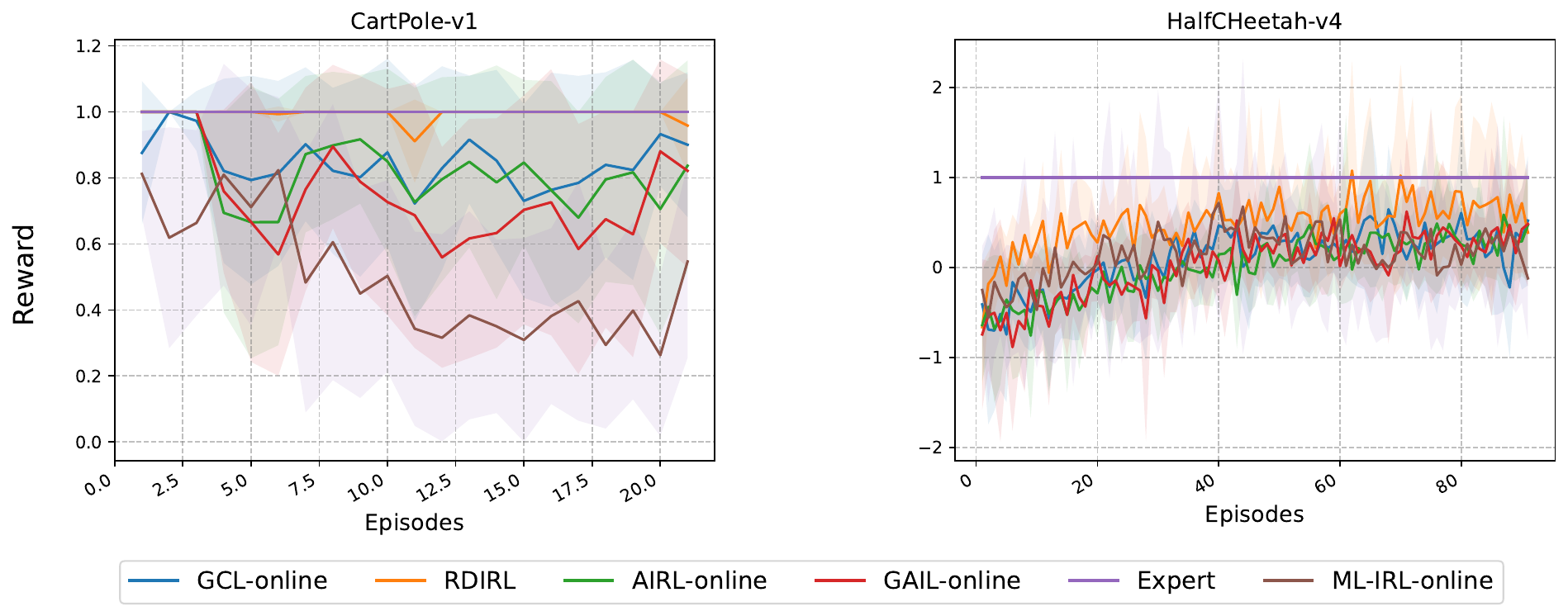}

\caption{Learning curves for RDIRL and online adaptation methods.}
\label{fig:results_online}
\end{figure}

As illustrated in Figure~\ref{fig:results_online}, our proposed method consistently outperforms the online-adapted baselines. Furthermore, the online adaptation does not significantly improve the performance of the original methods. In the case of Cartpole, it even leads to notable performance degradation and increased instability compared to both the original baselines (\gls{gail}, \gls{airl},\gls{mlirl}, \gls{gcl}) and our approach, as shown in Table~\ref{tab:scores_online}. These results highlight the advantage of our recursive optimization framework in producing more stable and accurate reward functions over naive online adaptation.

\begin{table}[ht]
\centering
\caption{Comparison of mean reward values for different Gym environments and online adapted methods.}
\label{tab:scores_online}
\setlength{\tabcolsep}{3.6pt}
\begin{tabular}{p{3.5cm}|p{2.7cm}|p{2.7cm}}
\toprule
Methods & CartPole & HalfCheetah-v4 \\ 
\midrule
GAIL                           & $0.934\pm0.058$      & $-0.521\pm1.15$     \\
GCL                            & $0.92\pm0.09$        & $-0.226\pm1.27$     \\
AIRL                           & $0.953\pm0.069$      & $-0.54\pm1.11$      \\
GAIL-Online                    & $0.74\pm0.29$        & $0.02\pm0.51$       \\
GCL-Online                     & $0.84\pm0.25$        & $0.1\pm0.53$        \\
AIRL-Online                    & $0.81\pm0.26$        & $0.01\pm0.49$       \\
ML-IRL-Online                  & $0.49\pm0.29$        & $0.14\pm0.75$       \\
\midrule
RDIRL (ours)                   & $\bm{0.99\pm0.13}$   & $\bm{0.49\pm0.59}$  \\
\bottomrule
\end{tabular}
\end{table}

\end{document}